\pdfoutput=1

\documentclass[11pt]{article}

\usepackage{EMNLP2022}

\usepackage{times}
\usepackage{latexsym}

\usepackage[T2A,T1]{fontenc}

\usepackage[utf8]{inputenc}
\usepackage[russian,english]{babel}

\usepackage{microtype}

\usepackage{inconsolata}

\usepackage{booktabs}
\usepackage{multirow}
\usepackage{graphicx}
\usepackage[multiple]{footmisc}

\title{{CUNI} Systems for the {WMT}~22 Czech-Ukrainian Translation Task}

\author{Martin Popel\footnotemark[1] \quad Jindřich Libovický\footnotemark[1] \quad Jindřich Helcl\footnotemark[1] \\
  Charles University, Faculty of Mathematics and Physics \\
  Institute of Formal and Applied Linguistics \\
  Malostranské náměstí 25, 118~00 Prague, Czech Republic \\
  \texttt{\{popel,libovicky,helcl\}@ufal.mff.cuni.cz}}

\begin{document}
\maketitle
\begin{abstract}
\let\thefootnote\relax\footnotetext{$^\ast$ The author order was determined by a coin toss.}We present Charles University submissions to the {WMT}~22 General
Translation Shared Task on Czech-Ukrainian and Ukrainian-Czech machine translation. We present two constrained submissions based on block back-translation and tagged back-translation and experiment with rule-based
romanization of Ukrainian. Our results show that the romanization only
has a minor effect on the translation quality.
Further, we describe Charles Translator,
a system that was developed in March 2022 as a response to the migration
from Ukraine to the Czech Republic. Compared to our constrained systems,
it did not use the romanization and used some proprietary data sources.
\end{abstract}

\section{Introduction}

How fast can the machine translation (MT) community react to a sudden need of
a high-quality MT system which was previously under low demand?
This question motivated the new task at the WMT this year, which is Czech-Ukrainian translation.

Both languages belong to the Slavic language family (Czech is western Slavic, Ukrainian is eastern Slavic), and share some lexical and structural characteristics. Unlike Czech, which uses the Latin script, Ukrainian uses its variant of the Cyrillic alphabet.

We submit three systems to the WMT~22 General Translation Shared Task for this language pair in each translation direction. The first system, 
\textsc{CUNI-JL-JH}, implemented in Marian \citep{junczys-dowmunt-etal-2018-marian}, uses tagged back-translation and is a result of our experiments with romanization of Ukrainian. 
Our second system,
\textsc{CUNI-Transformer}, implemented in Tensor2Tensor \citep{vaswani-etal-2018-tensor2tensor}, uses block back-translation.
Finally, we submit an unconstrained system, \textsc{Charles Translator}, implemented in Tensor2Tensor, which has been developed in spring 2022 as a response to the  crisis caused by the Russian invasion of Ukraine and the following migration wave.



\section{Constrained WMT Submissions}

We submitted two systems in each translation direction that use the same parallel and monolingual data, but different techniques and different toolkits. This section first describes the shared data processing steps and then the specifics of each of the submissions in separate subsections.

\subsection{Training Data}

We use all parallel data allowed in the constrained task, along with 50 million Czech and 58 million Ukrainian sentences of monolingual data. In the following paragraphs we describe the data cleaning steps when preparing the training data.
We further experiment with romanization of the Ukrainian Cyrillic alphabet and with artificial noising of the data.

\paragraph{Parallel data.}
The data for the constrained translation task consist of OPUS corpora \citep{tiedemann-2012-parallel} that have a Czech-Ukrainian part, WikiMatrix \citep{schwenk-etal-2021-wikimatrix} and the ELRC EU acts in Ukrainian.\footnote{
 \href{https://elrc-share.eu/repository/browse/eu-acts-in-ukrainian/71205868ae7011ec9c1a00155d026706d86232eb1bba43b691bdb6e8a8ec3ccf/}{https://elrc-share.eu/repository/search/?q=EU+acts+in+Ukrainian}
}

We clean the parallel data using rule-based filtering in the following way:
\begin{enumerate}
    \item Filter out non-printable and malformed UTF-8 characters.

    \item Detect language using FastText \citep{grave-etal-2018-learning}, only keep Czech and
        Ukrainian sentences on their respective source/target sides.

    \item Only keep sentence pairs with character length ratio between $0.67$
        and $1.5$ if longer than 10 characters.

    \item Apply hand-crafted regular expressions to filter out the frequent
        errors, such that the system does not attempt to translate e-mail
        addresses, currencies, etc. In addition, regular expressions check
        translations of names of Czech\footnote{\href{https://uk.wikipedia.org/wiki/\%D0\%9C\%D1\%96\%D1\%81\%D1\%82\%D0\%B0\_\%D0\%A7\%D0\%B5\%D1\%85\%D1\%96\%D1\%97}{https://uk.wikipedia.org/wiki/\foreignlanguage{russian}{Міста\_Чехії}}
}
        and Ukrainian\footnote{
\href{https://cs.wikipedia.org/wiki/Seznam_m\%C4\%9Bst_na_Ukrajin\%C4\%9B}{https://cs.wikipedia.org/wiki/Seznam\_měst\_na\_Ukrajině}
}
        municipalities downloaded from Wikipedia.
\end{enumerate}

We omit steps 2 and 3 for the XLEnt corpus, which seems to be very clean and consist
of short phrases (likely to get misclassified for language).

The sizes of the used parallel data sources before and after cleaning are presented in Table~\ref{tab:parallel}.

\begin{table}
\footnotesize\centering
\begin{tabular}{lrr}
\toprule
Source          & Original & Filtered \\
\midrule
bible-uedin     &      8 k  &     8 k \\
CCMatrix        &  3,992 k  & 3,884 k \\
EUbookshop      &      2 k  &     1 k \\
GNOME           &      150  &     81 \\
KDE4            &    134 k  &    64 k \\
MultiCCAligned  &  1,607 k  & 1,199 k \\
MultiParaCrawl  &  1,773 k  & 1,606 k \\
OpenSubtitles   &    731 k  &   273 k \\
QED             &    161 k  &   138 k \\
Tatoeba         &      3 k  &     2 k \\
TED2020         &    115 k  &   106 k \\
Ubuntu          &     0.2k  &    0.2k \\
wikimedia       &      2 k  &     2 k \\
XLEnt           &    695 k  &   695 k \\ \midrule

WikiMatrix      &    105 k  &    99 k \\
ELRC EU Acts    &    130 k  &   108 k \\ \midrule

Total           &  9,457 k  & 8,186 k \\
\bottomrule
\end{tabular}

\caption{Sizes of parallel data sources (number of sentence pairs).}\label{tab:parallel}
\end{table}

\paragraph{Monolingual data.}
The overview of the monolingual data sources is in Table~\ref{tab:mono}.
For Czech, we use the Czech monolingual portion of the CzEng 2.0 corpus \citep{kocmi-et-al-2020-czeng}.
For Ukrainian, we used all resources, available for WMT, i.e., the NewsCrawl,
the Leipzig Corpora \citep{biemann2007leipzig}, UberText corpus
\citep{khaburska2019toward} and Legal Ukrainian Crawling by ELRC.
The Uber corpus and the Ukrainian Legal corpus are distributed tokenized with
removed punctuation. We automatically restored the punctuation and detokenized
the models using a lightweight Transformer model (\citealp{vaswani2017attention}; Base model with 3 layers, 8k
vocabulary) trained on the NewsCrawl corpus.

For Ukrainian, we only keep sentences shorter than 300 characters.
For Czech, we keep all sentence lengths from the CzEng corpus (up to 1400 characters).
For both languages, we remove non-printable and malformed UTF-8 characters.

\begin{table}
\footnotesize\centering
\begin{tabular}{llrr}
\toprule
\multicolumn{2}{l}{Source} & Original & Filtered \\ \midrule

Czech & CzEng 2.0       & & 50.6 M \\ \midrule

\multirow{5}{*}{Ukrainian}
      & NewsCrawl       & \hphantom0 2.3 M & \hphantom0 2.0 M \\
      & Leipzig Corpora & \hphantom0 9.0 M & \hphantom0 7.6 M \\
      & UberText Corpus &           47.9 M &           41.2 M \\
      & ELRC Legal      & \hphantom0 7.6 M & \hphantom0 7.2 M \\  \cmidrule{2-4}
      & Total           &           66.8 M &           58.1 M \\
\bottomrule
\end{tabular}

\caption{Monolingual data sizes in number of sentences before and after filtering.}\label{tab:mono}

\end{table}

\paragraph{Romanization.} We develop a reversible romanization than transcribes
between the Ukrainian and Czech alphabets.
For example, \textit{\foreignlanguage{russian}{Зараз у нас є 4-місячні миші}} is transcribed to
 \textit{Zaraz u nas je 4-misjačni myši}.
This way the model can better
exploit the lexical similarities between the two languages
 (e.g. \textit{\foreignlanguage{russian}{миші}} should be translated to Czech as \textit{myši}),
 while keeping all the necessary information to reconstruct the original Cyrillic text.
Note that the transcription of Cyrillic changes when changing the target language, 
 reflecting the phonology of that language
 (e.g. \textit{\foreignlanguage{russian}{ш}} transcribes to \textit{sh} in English, but \textit{š} in Czech).
We introduce special tags for words and characters that are written in Latin script found in Cyrillic text.
The romanization is specifically designed for Ukrainian
 (e.g. \textit{\foreignlanguage{russian}{и}} transcribes to \textit{y},
 not \textit{i} as would be the case in Russian),
 so its reversibility occasionally fails for Russian names.

\paragraph{Artificial noise.}
We apply synthetic noise on the source side that should simulate the
most frequent deviations from the standard orthography (missing capitalization,
lower- or upper-casing parts of the sentences, missing or additional
punctuation).

\vspace{\baselineskip}

All scripts for training data processing are available at
\url{https://github.com/ufal/uk-cs-data-scripts}. 
We use Flores 101 \citep{goyal-etal-2022-flores} development set for validation.

\subsection{Tagged-back-translation-based System \textsc{(CUNI-JL-JH)}}

The \textsc{CUNI-JL-JH} submission is a constrained system and uses the data described in the paragraphs above. We train the system in 3 iterations of
tagged back-translation \citep{caswell-etal-2019-tagged} with greedy decoding. 
Each iteration, we filter the back-translated data using Dual Cross-Entropy filtering \citep{junczys-dowmunt-2018-dual} when keeping
$40,930,735$ synthetic sentences,
i.e., 5$\times$ the size of clean authentic parallel data.

The first two back-translation iterations were done with the Cyrillic script on the Ukrainian side.
In the final back-translation iteration, we performed 
romanization and noising of the source side. 
We train three models with random initialization and submit the ensemble.

For all iterations, we used a Transformer Big model with tied embeddings and  a shared SentencePiece vocabulary size of $32$k (fitted on 5M randomly sampled sentences; with sampling at the training time, $\alpha$=0.1; \citealp{kudo-richardson-2018-sentencepiece}). We set the learning rate to $0.0003$ and use $8,000$ warm-up steps. We initialize the models randomly in each back-translation iteration.

For validation, we use greedy decoding. At test time, we decode with beam search with beam size of 4 and length normalization of 1.0
(estimated on validation data).

The system is implemented using Marian \citep{junczys-dowmunt-etal-2018-marian}.

\paragraph{Negative results.}
We experimented with Dual-Cross-Entropy filtering \citep{junczys-dowmunt-2018-dual}
for parallel data
selection and came to inconclusive results. Therefore, we used
all parallel data after rule-based filtering.\footnote{
 Note that we use Dual-Cross-Entropy for filtering the monolingual data,
  as described in the first paragraph of this section,
  but we have not done any experiments with keeping all the monolingual data.
}

Additionally, we experimented with MASS-style \citep{song2019mass}
pre-training using monolingual data only
and continue with training on parallel data. We were not able to find a hyper-parameter setting
where the pre-trained model would outperform the models trained from random initialization.
Therefore, we only use model trained from random initialization.

\subsection{Block back-translation System \textsc{(CUNI-Transformer)}}

The CUNI-Transformer submission is also constrained,
 trained on the same data as \textsc{CUNI-JL-JH}.
The system was trained in the same way as the sentence-level English-Czech CUNI-Transformer systems
 submitted to previous years of WMT shared tasks \citep{popel-2018-cuni,popel-2020-cuni,gebauer-etal-2021-cuni}.
It uses Block back-translation (BlockBT) \citep{popel-et-al:2020},
 where blocks of authentic (human-translated parallel) and
 synthetic (backtranslated) training data are not shuffled together,
 but checkpoint averaging is used to find the optimal ratio of 
 checkpoints from the authentic and synthetic blocks (usually 5:3).
The uk$\rightarrow$cs system was trained
 with a non-iterated BlockBT
 (i.e. cs-mono data was translated with an authentic-only trained baseline).
The cs$\rightarrow$uk was trained with two iterations of BlockBT
 (i.e. the uk-mono data was translated with the above mentioned uk$\rightarrow$cs non-iterated BlockBT system).
We had not enough time to train more iterations and apply noised training and romanization.
The system was implemented using Tensor2Tensor \citep{vaswani-etal-2018-tensor2tensor}.

\paragraph{Inline casing.}
We experimented with Inline casing (InCa) pre-processing in the cs$\rightarrow$uk direction.
The main idea is to lowercase all training data and
 insert special tags \texttt{<titlecase>} and \texttt{<all-uppercase>} before words in the respective case, so that the original casing can be reconstructed
 (with the exception of words like \textit{McDonald} or \textit{iPhone},
 which use different casing patterns than all-lowercase, all-uppercase and titlecase).
We improved this approach by remembering the most frequent casing variant of each (lowercased) word in the training data.
The most frequent variant does not need to be prefixed with any tag,
 which makes the length of training sequences shorter.
We also introduced a third tag \texttt{<all-lowercase>} for encoding all-lowercased words
 whose most frequent variant is different.
For example, if the InCa vocabulary includes only two items: \textit{iPhone} and \textit{GB},
 sentence \textit{My iPhone 64GB and iPod 64 GB or 32 gb} will be encoded as
 \textit{\texttt{<titlecase>} my iphone \texttt{<all-uppercase>} 64gb and iPod 64 gb or 32 \textit{<all-lowercase>} gb}.
Note that \textit{iPod} was kept in the original case because it was not included in the InCa vocabulary and it does not match any of the three ``regular'' casing patterns.
We applied InCa on both the source and target side
 and experimented with training the InCa vocabulary on the authentic data only
 or on authentic plus synthetic (monolingual backtranslated).

Inline casing showed promising results in preliminary experiments (without backtranslation),
 especially when combined with romanization and artificial noise in training.
Unfortunately, we had not enough time to train the backtranslated model long enough,
 so we submitted it only as a contrastive run and plan to explore it more in future.

\section{Charles Translator for Ukraine}

Charles Translator for Ukraine is a free Czech-Ukrainian online translation
service available for public at \url{https://translator.cuni.cz} and as an
Android app.  It was developed at Charles University in March 2022 to help
refugees from Ukraine by narrowing the communication gap between them and other
people in Czechia. 
Similarly to \textsc{CUNI-Transformer},
 it is based on Transformer and iterated Block back-translation \citep{popel-et-al:2020}.
The training used source-side artificial noising,
 but no romanization and no inline casing.
It was trained on most (but not all) of the
training data provided by WMT plus about one million uk-cs sentences from the
InterCorp v14 corpus \citep{cermak-rosen:2012,kotsyba:2022}, so this submission
is unconstrained.

\section{Results}

In this section, we report BLEU scores on the Flores 101 development set that
we used to make our decisions about the system development and the final
automatic scores. Note that the validation set is very different from the test
set. The validation set consists of clean and rather complicated sentences from
Wikipedia articles, whereas the WMT~22 test set is noisy user-generated text
from the logs of the production deployment of Charles Translator.\footnote{
 The test set only contains sentences from users who provided their consent
  for this usage and the sentences were pseudonymized.
}

\paragraph{Tagged BT systems.}
Table~\ref{tab:first_bt} shows validation BLEU scores from the first three
iterations of back-translation.
The second and third iteration did not bring substantial improvements,
 so we decided not to further iterate.

\begin{table}
    \centering\footnotesize
    \begin{tabular}{lcc}
    \toprule
    Model & cs$\rightarrow$uk & uk$\rightarrow$cs \\ \midrule

    Authentic only          & 20.91 & 22.95  \\
    BT iteration 1          & 21.69 & 23.70 \\
    BT iteration 2          & 21.87 & 23.98 \\
    BT iteration 3 (seed 1) & 21.53 & 23.76 \\

     \bottomrule
    \end{tabular}
    \caption{Validation BLEU scores for the first two iterations of BT for the tagged BT systems.}\label{tab:first_bt}
\end{table}

Table~\ref{tab:ensemble} shows validation BLEU scores from the last (third) BT
iteration -- three independently trained systems and their ensembles, and the
Cyrillic and romanized versions of the data. In general, ensembling only brings
a small improvement. Romanization does not bring a significant difference
compared to using the Cyrillic script. In the Czech-to-Ukrainian direction, the
best system was the ensemble of the romanized systems. However, in the
Ukrainian-to-Czech direction, the best system was one of the Cyrillic systems
that used accidentally 3 times higher batch size than the remaining ones. This
result suggests that the batch size has a much stronger effect than most of the
techniques that we experimented with and that we might have reached better
results if we opted for higher batch size.

\begin{table}
    \centering\footnotesize
    \begin{tabular}{llcc}
    \toprule
    \multicolumn{2}{l}{Model} & cs$\rightarrow$uk & uk$\rightarrow$cs \\ \midrule

    \multirow{4}{*}{\rotatebox{90}{Cyrillic~~}}
       & Seed 1   & 21.53 & 23.76 \\
       & Seed 2   & 22.28 & \bf 25.10 \\
       & Seed 3   & 21.96 & 24.39 \\ \cmidrule{2-4}
       & Ensemble & 22.45 & 24.86 \\ \midrule

    \multirow{4}{*}{\rotatebox{90}{Romanized~}}
       & Seed 1   & 21.42 & 23.99 \\
       & Seed 2   & 21.76 & 23.91 \\
       & Seed 3   & 22.37 & 24.18 \\ \cmidrule{2-4}
       & Ensemble & \bf 22.62 & 24.22 \\
    \bottomrule
    \end{tabular}

    \caption{Validation BLEU scores for the last (i.e., the third) iteration of BT comparing
    romanized and original script.}\label{tab:ensemble}

\end{table}


\begin{table*}
    \centering \footnotesize
    \begin{tabular}{l ccc ccc}
    \toprule
     & \multicolumn{3}{c}{cs$\rightarrow$uk} & \multicolumn{3}{c}{uk$\rightarrow$cs} \\ \cmidrule(lr){2-4} \cmidrule(lr){5-7}
    System & BLEU & chrF & COMET & BLEU & chrF & COMET \\ \midrule

    Best constrained (HuaweiTSC/AMU)         & 36.0 & 62.6 & 0.994   & 37.0 & 60.7 &  1.048 \\
    CUNI-Transformer                         & 35.0 & 61.6 & 0.873   & 35.8 & 59.0 &  0.885 \\
    CUNI-JL-JH                               & 34.8 & 61.6 & 0.900   & 35.1 & 58.7 &  0.890 \\ \midrule

    Best unconstrained (Lan-Bridge/Online-B) & 38.1 & 64.0 & 0.942   & 36.5 & 60.4 &  0.965 \\
    Charles Translator                       & 34.3 & 61.5 & 0.908   & 35.9 & 59.0 &  0.901 \\

    \bottomrule
    \end{tabular}
    \caption{Final automatic results on the WTM22 test data compared to the best overall score achieved in each metric.
    }
    \label{tab:final}
\end{table*}

\paragraph{Results on WMT test.}

Automatic evaluation on the WMT22 test set is presented in Table~\ref{tab:final}.
Both the constrained systems and Charles Translator show comparable results.
The tagged BT system reaches a slightly higher COMET score than the Block BT system,
however, Czech-Ukrainian was not in the training data of the COMET score, which make
the score unreliable for this particular language pair.
For Czech-to-Ukrainian, Charles Translator reaches a slightly higher COMET score
 and slightly lower BLEU and chrF scores than both the constrained systems,
 but we do not consider such small differences of automatic metrics relevant.

\section{Conclusions}

We presented Charles University submissions to the {WMT}~22 General
Translation Shared Task on Czech-Ukrainian and Ukrainian-Czech machine translation. We present two constrained submissions based on block back-translation and tagged back-translation and experiment with rule-based
romanization of Ukrainian. 
Further, we describe Charles Translator,
a system that was developed in March 2022 as a response to the migration
from Ukraine to the Czech Republic. Compared to our constrained systems,
it did not use the romanization and used some proprietary data sources.

Our results show that the romanization only
has a minor effect on the translation quality, compared to machine-learning
aspects that affect translation quality. Block back-translation seems
to deliver slightly better results that tagged back-translation, however
the differences are only small.

%
\section*{Acknowledgements}

This work has been supported
 by the Ministry of Education, Youth and Sports of the Czech Republic, Project No. LM2018101 LINDAT/CLARIAH-CZ,
 by the Czech Science Foundation (GACR) grant 20-16819X (LUSyD),
 and by the European Commission via its
  Horizon 2020 research and innovation programme no. 870930 (WELCOME),
  Horizon Europe Innovation programme no. 101070350 (HPLT).

\bibliography{anthology,custom}
\bibliographystyle{acl_natbib}


\end{document}